\documentclass[12pt]{alt2024}

\title[Adversarial Contextual Bandits Go Kernelized]{Adversarial Contextual Bandits Go Kernelized}
\usepackage{times}

\usepackage[utf8]{inputenc} 
\usepackage[T1]{fontenc}    
\usepackage{hyperref}       
\usepackage{url}            
\usepackage{booktabs}       
\usepackage{amsfonts}       
\usepackage{nicefrac}       
\usepackage{microtype}      
\usepackage{xcolor}         

\usepackage{algorithm}

\usepackage{xspace}
\usepackage{comment}
\usepackage{selectp}
\usepackage{setspace,lipsum}

\usepackage[utf8]{inputenc}
\usepackage{amsmath}
\usepackage{amsfonts}
\usepackage{amssymb}

\usepackage{bookmark}
\usepackage{comment}
\usepackage{amsmath}

\usepackage{amssymb}
\usepackage{tikz}
\usetikzlibrary{arrows,decorations.markings}
\usepackage{multirow}
\usepackage{multicol}
\usepackage{enumitem}
\usepackage{hyperref}
\usepackage{times}
\usepackage[noend]{algpseudocode}

\usepackage{graphicx}

\usepackage{times}

 \newcommand{\Hil}{\mathcal{H}}

\newcommand{\X}{\mathcal{X}}
\newcommand{\F}{\mathcal{F}}

\newcommand{\real}{\mathbb{R}}

\newcommand{\Dw}{\mathcal{D}}

\newcommand{\BB}{\mathcal{B}}
\newcommand{\DD}{\mathcal{D}}
\newcommand{\OO}{\mathcal{O}}
\newcommand{\tOO}{\wt{\OO}}
\newcommand{\trace}[1]{\mbox{tr}\left(#1\right)}
\newcommand{\II}[1]{\mathbb{I}_{\left\{#1\right\}}}
\newcommand{\PP}[1]{\mathbb{P}\left[#1\right]}

\newcommand{\EE}[1]{\mathbb{E}\left[#1\right]}

\newcommand{\EEtb}[1]{\mathbb{E}_t\bigl[#1\bigr]}

\newcommand{\PPt}[1]{\mathbb{P}_t\left[#1\right]}
\newcommand{\EEt}[1]{\mathbb{E}_t\left[#1\right]}

\newcommand{\EEcct}[2]{\mathbb{E}_t\left[\left.#1\right|#2\right]}

\def\argmin{\mathop{\mbox{ arg\,min}}}

\newcommand{\ra}{\rightarrow}

\newcommand{\siprod}[2]{\langle#1,#2\rangle}
\newcommand{\iprod}[2]{\left\langle#1,#2\right\rangle}
\newcommand{\biprod}[2]{\bigl\langle#1,#2\bigr\rangle}

\newcommand{\norm}[1]{\left\|#1\right\|}
\newcommand{\bnorm}[1]{\bigl\|#1\bigr\|}

\newcommand{\opnorm}[1]{\norm{#1}_{\text{op}}}

\newcommand{\ev}[1]{\left\{#1\right\}}
\newcommand{\pa}[1]{\left(#1\right)}
\newcommand{\bpa}[1]{\bigl(#1\bigr)}

\newcommand{\wh}{\widehat}
\newcommand{\wt}{\widetilde}

\newcommand{\loss}{\ell}
\newcommand{\hloss}{\wh{\loss}}

\newcommand{\tX}{X_0}

\newcommand{\Sp}{\Sigma^+}
\newcommand{\Spt}{\pa{\Sigma^{+}}^2}
\newcommand{\hSp}{\widehat{\Sigma}^+}
 \newcommand{\hSigma}{\widehat{\Sigma}} 
 \newcommand{\A}{\mathcal{A}}

\newcommand{\hS}{\wh{\Sigma}}

\definecolor{PalePurp}{rgb}{0.66,0.57,0.66}

\newcommand{\hL}{\wh{L}}

\newcommand{\kerlinexp}{\textsc{KernelFTRL}\xspace}
\newcommand{\lossestimate}{\textsc{KGRLossEstimate}\xspace}
\newcommand{\KGR}{\textsc{KGR}\xspace}

\def\qed{\hfill$\Box$\medskip}

\altauthor{%
 \Name{Gergely Neu} \Email{gergely.neu@gmail.com}\\
 \addr Universitat Pompeu Fabra
 \AND
 \Name{Julia Olkhovskaya} \Email{julia.olkhovskaya@gmail.com}\\
 \addr TU Delft%
  \AND
 \Name{Sattar Vakili} \Email{sattar.vakili@mtkresearch.com}\\
 \addr MediaTek Research%
}

\def\Hc{\mathcal{H}}

\newtheorem{assumption}{Assumption}

\begin{document}
\maketitle
\begin{abstract}
We study a generalization of the problem of online learning in  adversarial  linear contextual bandits by incorporating loss functions that belong to a reproducing kernel Hilbert space, which allows for a more flexible modeling of complex decision-making scenarios. We propose a computationally efficient algorithm that makes use of a new optimistically biased estimator for the loss functions and achieves near-optimal regret guarantees under a variety of eigenvalue decay assumptions made on the underlying kernel. Specifically, under the assumption of polynomial eigendecay with exponent~$c>1$, the regret is $\tOO(KT^{\frac{1}{2}\pa{1+\frac{1}{c}}})$, where $T$ denotes the number of rounds and $K$ the number of actions. Furthermore, when the eigendecay follows an exponential pattern, we achieve an even tighter regret bound of $\tOO(\sqrt{T})$. These rates match the lower bounds in all special cases where lower bounds are known at all, and match the best known upper bounds available for the more well-studied stochastic counterpart of our problem.

\end{abstract}

\section{Introduction}

	In the domain of sequential decision-making, the framework of contextual bandits has emerged as an important tool for modeling interactions between a learner and environment in a sequence of rounds. Within each such round, the learner observes a context and subsequently selects an action and incurs a loss.  The objective of the learner in this iterative process is to minimize her cumulative losses over a sequence of rounds. This  model has been employed in a large variety of applications, including medical treatments  \citep{Tewari2017}, the domain of personalized recommendations  \citep{beygelzimer2011contextual}, and online advertising \citep{chu2011contextual}.
	
  One of the main challenges of the contextual bandit problem is that the partial observations made about the losses handed out by the environment must be generalized efficiently to a possibly infinite set of contexts that are yet to be encountered in future decision-making rounds. One possible way to address this challenge is by making suitable assumptions about the structure of the losses.  One particularly well-studied model is that of  \emph{linear contextual bandit}, where the losses are assumed to be linear in some known low-dimensional representation of the contexts. In the most broadly considered version of this setup, the sequence of contexts is completely arbitrary and the losses are determined by fixed linear functions. Advancements in this model have been made in a range of works, including  \cite{chu2011contextual, abbasi2011improved, li2019nearly, foster2020adapting}. 

  This model has been successfully generalized to deal with non-linear loss functions that belong to reproducing kernel Hilbert spaces. This assumption is broadly applicable, as the RKHS associated with commonly used kernels has the capacity to approximate nearly all continuous functions on compact subsets of $\mathbb{R}^d$ \citep{SS02,RW06}.  Viewed through the lens of kernel maps, this setting represents an extreme extension of the parametric linear bandit setting mentioned above, where the contexts can be represented in infinite-dimensional vector spaces.  Works like \citep{srinivas2009gaussian,VKMFC13,chowdhury2017kernelized} have provided efficient algorithms with strong performance guarantees for contextual bandits with such nonlinear loss functions that remain fixed throughout the online learning process.

	
The primary focus of this paper lies in a distinct model known as the adversarial contextual bandit. In this setup, we assume that the context is drawn from a fixed distribution, and losses are chosen by a potentially adaptive adversary. For this setting, the simplest approach is to make use of a  finite class of policies that map contexts to actions, as done by the classic \textsc{Exp4} algorithm of \citet{auer2002nonstochastic}. An alternative to this line of work takes inspiration from the stochastic linear contextual bandit literature, and models the losses as linear functions of some known finite-dimensional feature map~\citep{NO20, liu2023bypassing}.

	
Our principal contribution is extending the understanding of the adversarial linear contextual bandit model to work with a large class of nonlinear loss functions. To enhance model flexibility, we consider the setting where the sequence of loss functions drawn by the adversary belong to a fixed and known RKHS. Within this framework, we establish a regret bound of $\tOO(KT^{1/2(1+1/c)})$ for loss functions characterized by polynomial eigendecay ($\mu_i  = \OO(i^{-c})$) and a $\tOO(K\sqrt{T})$ bound for those exhibiting exponential eigendecay ($\mu_i =\OO( e^{-ci})$). 
These conditions are well-studied in the broader literature on learning with kernels, and in particular our results align with the lower bounds established for kernelized bandits with adversarial losses by \cite{CPB19}, and match the best known upper bounds in the stochastic version of our problem by \citet{VKMFC13}. 

At a high level, our approach is based on the regret decomposition idea of \citet{NO20} originally proposed for finite-dimensional linear bandits: we place a suitably chosen online learning algorithm in each context $x$, and feed each algorithm with a suitably chosen estimator for the loss functions that allows generalization across different contexts. Our key technical contribution is the construction of an optimistically biased loss estimator that can be effectively computed via a kernelized version of the Matrix Geometric Resampling estimator proposed for finite-dimensional linear losses by \citet{NO20}. The optimistic bias is achieved by adding a context-dependent exploration bonus to the standard estimator, in order to offset its potentially large positive bias that could otherwise be problematic to handle for a standard analysis. Another key component of our algorithm design is the now-classic log-barrier regularization function popularized in the online learning literature by \citet{foster2016learning}---see also the earlier works of \citet{DKJSD07,JKDG09,KB10,ACLR15,Chr16} and follow-ups by \citet{ALNS17,BCL18,WL18,LWZ18} that made use of the same regularizer. In our case, we use the special property of the log-barrier that it can appropriately handle loss functions in an FTRL scheme that are potentially unbounded (as will be the case with our estimators).

	The remainder of this paper is structured as follows. In the next section, we introduce the essential notation and definitions. Section~\ref{s:algo} presents our algorithm and provides its performance guarantees. Detailed proofs supporting our analysis can be found in Section~\ref{s:analysis}. We draw our conclusions in Section~\ref{s:discussion}, where we also delve into the implications of our results.
	
	\paragraph{Notation.} We let $\ell_2$  denote the space of square-summable sequences. For any two elements $v,w\in\ell_2$, we use $\iprod{w}{v}$ to denote the standard $\ell_2$ inner product $\sum_{i=1}^\infty w_iu_i$, and we define the $\ell_2$ norm of $v$ as $\norm{v}_2 = \sqrt{\iprod{v}{v}}$. The tensor product of $v$ and $w$ is denoted by $v\otimes w$, and is defined as the operator that acts on elements $u$ of $\ell_2$ as $\pa{v\otimes w}u = v \iprod{w}{u}$. For a positive definite operator $B$ on $\ell_2$, we define $\norm{v}_{B} = \sqrt{\iprod{v}{Bv}}$, and its trace as $\trace{B}=\sum_{i} \norm{e_i}_B^2$, where $e_i$ is the $i$th canonical basis vector in $\ell_2$. In the context of sequential-decision making problems, we will use $\F_t$ to denote the interaction history between the learner and the environment, and use the shorthand notations 
	$\EEt{ \cdot} = \EE{ \cdot| \mathcal{F}_{t-1}}$ and $\PPt{ \cdot} = \PP{ \cdot| \mathcal{F}_{t-1}}$.
	
\label{sec:prelim}
\section{Preliminaries}
We now introduce our learning setting and the assumptions that we make about the loss functions. 

\subsection{Adversarial contextual bandits}
We investigate a sequential interaction scheme between a learner and its environment, where the subsequent steps are iteratively executed over a fixed number of rounds $t=1,2,\dots,T$:
\begin{enumerate}[noitemsep]
        \item The environment draws the context vector 
	$X_t\in\real^d$ from the context distribution $\Dw$, and reveals it to the learner;
	\item Independently of the context $X_t$, the environment chooses a loss function $\ell_t:\X\times\A\ra[0,1]$;
	\item Based on $X_t$ 
	and possibly some randomness, the learner chooses action $A_t\in[K]$;
	\item The learner incurs and observes loss $\ell_{t}(X_t,A_t)$.
\end{enumerate}
The primary objective of the learner is to strategically choose actions to minimize its cumulative loss. It is important to note that we refrain from making any statistical assumptions about the sequence of losses. In fact, we allow these losses to depend on the entire historical interaction, making it impractical for the learner to aim for a loss level as low as that of the best sequence of actions. A more realistic goal is to strive to match the performance of the best fixed policy that maps contexts to actions. To formalize this objective, the learner considers the set $\Pi$, which contains all policies  $\pi:\real^d\ra[K]$, and seeks to minimize its total expected regret,  which is formally defined as
\[
R_T = \sup_{\pi\in\Pi} \EE{\sum_{t=1}^T \bpa{\ell_{t}(X_t,A_t) - \ell_{t}(X_t,\pi(X_t))}}.
\]
Here, the expectation is taken over the randomness injected by the learner, as well as the 
sequence of random contexts. 
It is easy to show  that the optimal policy $\pi_T^*$, which serves as the benchmark for the learner's performance, is defined by the following rule:
\begin{equation}\label{best_policy}
	\pi_T^*(x) = \argmin_a \sum_{t=1}^T \loss_{t}(x,a),\qquad\qquad \forall x\in\real^d.
\end{equation}

\subsection{RKHS loss functions}\label{sec:kernels}
Throughout the paper, we will make the assumption that the loss functions $\ell_{t}(\cdot,a)$ belong to a known reproducing kernel Hilbert space (RKHS) for each $t,a$. Specifically, we will suppose that the space of contexts $\X\subseteq \real^d$, and we are given a positive definite kernel $\kappa: \mathcal{X} \times \mathcal{X} \rightarrow \mathbb{R}$. We let  $\Hil_\kappa\subseteq \real^{\X}$ be the RKHS induced by $\kappa$. Without
loss of generality, we assume $\kappa(x,x)\le 1$ for all $x\in\X$. The inner product and norm of $\Hil_\kappa$ are represented by $\langle\cdot, \cdot\rangle_{\Hil_\kappa}: \Hil_\kappa \times \Hil_\kappa \rightarrow \mathbb{R}$ and $\|\cdot\|_{\Hil_\kappa}: \Hc_\kappa \rightarrow \mathbb{R}$, respectively. Mercer's theorem implies that, under certain mild conditions, $\kappa$ can be represented using an infinite-dimensional feature map:
\begin{eqnarray}\label{eq:Mercer}
\kappa(x,x')=\sum_{j=1}^{\infty}\mu_j\psi_j(x)\psi_j(x'),
\end{eqnarray}
where $\mu_j\in\real_+$ are the Mercer eigenvalues and $\psi_j\in\Hil_\kappa$ are the corresponding eigenfunctions, and $\sqrt{\mu_j}\psi_j$ form an orthonormal basis of $\Hil_\kappa$. 
Using this basis, any $h \in \Hil_\kappa$ can be represented through the real-valued square summable sequence $\pa{w_j}_{j=1}^\infty \in \ell_2$ as
\begin{equation*}
	h = \sum_{j=1}^{\infty} w_j \sqrt{\mu_j}\psi_j,
\end{equation*}
where $\norm{h}^2_{\Hil_\kappa} = \sum_{j=1}^{\infty}w_j^2$. 
A formal statement and the details can be found in Appendix~\ref{appx:mercer}. 
We will use the notation $\varphi_i(x) = \sqrt{\mu_i} \psi_{i}(x)$ and use $\varphi(x) = \pa{  \varphi_i(x)}_{i=1}^{\infty} \in \ell_2$ to denote the representation of context $x$ in $\ell_2$ induced by Mercer's theorem. An important implication of Mercer's theorem that we will repeatedly use is that $\iprod{\varphi(x)}{\varphi(x')} = \kappa(x,x')$ holds for all $x,x'\in\X$.

Attempting to obtain a sublinear regret bound  without any assumptions regarding the regularity of the loss function would be an arduous task. In this paper, we will impose such regularity conditions by making assumptions about the Mercer eigenvalues of the kernel $\kappa$.
\begin{assumption}\label{a:eigendecay}
	We assume that  the Mercer eigenvalues  $\{\mu_j\}_{j\ge 1}$  of  the kernel $\kappa$ over $\X$   are ordered as $\mu_1 \ge \mu_2 \ge \dots$, and are such that  they meet one of the following two eigenvalue decay profiles for some constants $g >0, c > 0$:
	\begin{itemize}
		\item $(g,c)$-exponential decay: for all $j\in \mathbb{N}$, we  have $\mu_{j}\le g e^{- cj}$.
		\item $(g,c)$-polynomial decay with $c > 1$: for all $j\in \mathbb{N}$, we  have $\mu_j \le g j^{-c}$. 
	\end{itemize}
\end{assumption}
As an alternative way to measure the decay rate of the kernel $\kappa$, we will also define the following quantity for each $\varepsilon>0$: 
\[
m(\varepsilon) = \min\ev{m\in\mathbb{N}: \textstyle\sum_{j=m+1}^\infty \mu_j \le \varepsilon}.
\]
It is easy to see that or kernels that satisfy the exponential decay condition, $m(\varepsilon) = \OO\pa{\log(g/(c\varepsilon))/c }$ and  for kernels that satisfy the polynomial decay condition, 
 $m(\varepsilon) = \OO\pa{\pa{(c-1)\varepsilon/g }^{1/(1-c)}}$.
Many practically used kernels are consistent with    Assumption~\ref{a:eigendecay}. For instance, the squared exponential kernel satisfies the exponential decay condition with $c = 1/d$, and the Mat\'ern kernel with smoothness parameter $v>2$ satisfies the polynomial decay condition with $c = 1 + 2v/d$. We refer to \cite{SMKF08} and the discussion in \cite{yang2020function} for proofs of these facts and further examples.   

Now we can precisely state our assumptions on the loss functions and the contexts.  We will suppose the context distribution is supported on the bounded set $\X \subset \real^d$ with each $x\in \X$ satisfying 
$\norm{\varphi(x)}_{2} \le 1 $ and that the loss function is bounded by one in absolute value:
$\big|\loss_{t}(x,a) \big| \le 1$ for all $t$, $a$ and all $x \in \X$. Furthermore, we will suppose that the loss function satisfies $\ell_t(\cdot,a)\in \Hil_\kappa$ and in particular that it can be written as $\ell_t(x,a) = \iprod{f_{t,a}}{\varphi(x)}$ for some $f_{t,a}\in\ell_2$ that satisfies $\norm{f_{t,a}}_{2} \le 1$ for all $t, a$.

\section{Algorithm and main result}\label{s:algo}
We now present our algorithm which is based on a regret-decomposition approach first proposed by \citet{NO20} for 
finite-dimensional linear contextual bandits. The core idea of this method is to instantiate an online learning 
algorithm in every context $x\in\X$ and feed it with an appropriately designed estimator of the loss function that 
allows generalization across different contexts. Concretely, we will run an instance of the 
standard Follow-the-Regularized-Leader (FTRL) algorithm with log-barrier regularization (as popularized in online 
learning by \citealp{foster2016learning}) as the online learning method, and derive a new loss estimator based on the 
Matrix Geometric Resampling procedure proposed by \citet{NO20} along with an optimistic exploration idea that is novel 
within this context. While the algorithm formally needs to calculate its policies and loss estimates that are valid on the whole context-action space, we will show that it can be implemented efficiently by querying the policy and the estimates only in the contexts encountered in runtime.
To preserve readability in this section, we present a relatively abstract version of our algorithm first without worrying about implementability, and defer a fully detailed operational description to Appendix~\ref{appx:MGR}.

 
We start by describing the algorithm (that we call \kerlinexp) for a generic choice of loss estimators 
$\hloss_t:\X\times\A\ra \real$ whose specifics will be described shortly. Letting  $\widehat{L}_{t}(x,a) = \sum_{\tau=1}^{t} \hloss_{\tau}(x,a)$ denote the 
cumulative sum of the estimated losses, our algorithm calculates its policy $\pi_t:\X\ra \Delta_{\A}$ by solving the following optimization problem in 
each round $t$:
\[
\pi_{t}(\cdot|X_t) = \argmin_{p \in \Delta(\A)}\pa{\Psi(p) + \eta \sum_{a}p_a \widehat{L}_{t-1}(X_t,a) }.
\]
Here, $\eta > 0$ serves as a learning-rate parameter and $\Psi(p) = \sum_{a\in \A} \ln\pa{\frac{1}{p_a}}$. Note that 
the algorithm only has to compute the distribution $\pi_t(\cdot|X_t)$ \emph{locally at $X_t$}
which can be done efficiently as long as $\widehat{L}_{t-1}(X_t,a)$ can be efficiently computed for all actions $a$ \citep{foster2016learning}. We 
will show later that this condition holds true for our loss estimators.  We present this method as 
Algorithm~\ref{alg:kerlinexp3} below.

\begin{algorithm}
	\caption{\kerlinexp}
	\label{alg:kerlinexp3}
	\textbf{Parameters:} Learning rate $\eta>0$.\\
	\textbf{Initialization:} Set $\BB_t = \emptyset$.
	\\
	\textbf{For} $t = 1, \dots, T$, \textbf{repeat:}
	\begin{enumerate}[noitemsep]
		\item Observe $X_t$ and, for all $a$, set 
		\begin{equation}\label{eq:FTRL}
			\pi_{t}(\cdot|X_t) = \argmin_{p \in \Delta(\A)}\pa{\Psi(p) + \eta\sum_a p_a \widehat{L}_{t-1}(X_t,a) }, 
		\end{equation}
		\item draw $A_t$ from the policy $\pi_t(\cdot | X_t)$,
		\item observe the loss $\loss_t(X_t,A_t)$ and call Kernel Geometric Resampling to produce $\hloss_t$.
	\end{enumerate}
\end{algorithm}
To describe our loss estimator, we first introduce the following  operator on  $\ell_2$: 
\begin{equation}\label{cov_m}
	\Sigma_{t,a} = \EEt{\II{A_t = a} \pa{\varphi(X_t) \otimes \varphi(X_t)}}.
\end{equation}
Then, supposing for the sake of argument that $\Sigma_{t,a}$ is invertible  (which will not be necessary for our actual 
estimator), we can define the estimate 
\[  \widehat{f}_{t,a} = \Sigma_{t,a}^{-1}\varphi(X_t) \ell_{t,a} \II{A_t = a}, \]
which can be easily demonstrated to be unbiased:
\[ \EEt{ \wh{f}_{t,a} } = \EEt {\Sigma_{t,a}^{-1}\varphi(X_t) \ell_{t,a} \II{A_t = a} } = \EEt {\Sigma_{t,a}^{-1} 
\II{A_t = a} \varphi(X_t) \siprod{\varphi(X_t)}{f_{t,a}} } = f_{t,a}. \]
Here, we used that $\ell_{t}(X_t,a)  = \siprod{\varphi(X_t)}{f_{t,a}}$, and that 
$\varphi(X_t) \siprod{\varphi(X_t)}{f_{t,a}} = \pa{\varphi(X_t) \otimes \varphi(X_t)} f_{t,a}$ holds by 
definition of the tensor product.  Note that the dimension of the $\Hil_\kappa$ could be infinite (for example when $\kappa$ is 
the Gaussian kernel), so neither $\Sigma_{t,a} $ or $\widehat{f}_{t,a}$ can be computed explicitly. Another 
challenge is that, even in the case of a fixed dimension, the operator $\Sigma_{t,a}$ relies on the joint distribution 
of both the context $X_t$ and the action $A_t$, which exhibits a highly intricate structure. As a final note, the 
eigenvalues of $\Sigma_{t,a}$ can be arbitrary small, which may result in a loss estimator  of unbounded norm $\bnorm{\widehat{f}_{t,a}}_{2}$ even in the unlikely case that $\Sigma_{t,a}$ is invertible. 

To deal with the difficulties stated above, we propose an estimator derived by adapting the idea of Matrix Geometric 
Resampling (MGR) from \cite{NO20} to the kernel setting. The  estimator is efficiently computable, but requires 
sampling access to the context distribution $\mathcal{D}$. To deal with the bias of the standard MGR estimator, we also 
introduce a new element in our algorithm design: an \emph{optimistic exploration bonus} whose purpose is to make sure 
that the estimates are negatively biased which we will see to be beneficial to the analysis. The bonus for 
context-action pair $(x,a)$ added in round $t$ will be denoted by $b_t(x,a)$, and will be computed within the same 
procedure as the base loss estimates themselves. The procedure (which we call Kernel Geometric Resampling or KGR) is 
presented below:

\vspace{.35cm}
\makebox[\textwidth][c]{
	\fbox{
		\begin{minipage}{.9\textwidth}
			\textbf{Kernel Geometric Resampling}
			\vspace{.1cm}
			\hrule
			\vspace{.1cm}
			\textbf{Input:} Context $x, X_t$, data distribution $\Dw$, parameters $\beta, M$.\\
			\textbf{For $k = 1, \dots, M$, repeat}:
			\begin{enumerate}[noitemsep]
				\vspace{-2mm}
				\item Draw $X(k)\sim\Dw$ and $A(k) \sim \pi_t(\cdot|X(k))$,
				\item compute $q_{t,k,a}(x)= \iprod{\varphi(x)}{C_{k,a} \varphi(X_t)}$ and $b_{t,k,a}(x) = 
\beta \iprod{\varphi(x)}{C_{k,a} \varphi(x)}$, \\where $C_{k,a} = \prod_{j=1}^k\pa{I-  B_{j,a}}$ \\
				and $B_{k,a} =  \II{A(k)=a}\varphi(X(k))\otimes\varphi(X(k))$.
			\end{enumerate}
			\vspace{-2mm}
			\textbf{Return $q_{t}(x,a) =  \kappa(x, X_t) +  \sum_{k=1}^M q_{t,k,a}(x),$}
	
			$\	\qquad \quad b_{t}(x,a) = \kappa(x, x) + \sum_{k=1}^M b_{t,k,a}(x)$.
		\end{minipage}
	}
}\vspace{\baselineskip}
Then, the estimator of  $\ell_t(x,a)$ can be written as
$$ \widehat\ell_t(x,a)= q_{t}(x,a) \loss_t(X_t, a) \II{A_t = a} - b_{t}(x,a).$$ 
Notice that all operations performed by Kernel Geometric Resampling can be implemented by applying simple rank-one 
operators to elements of $\ell_2$,
so $ \wh\ell_t(x,a)$ can be computed without having to hold in memory $C_{k,a}$ and $B_{k,a}$, which can both be 
infinite-dimensional objects. In Appendix~\ref{appx:MGR}, we show that $q_{t}(x,a)$ and 
$b_t(x,a)$ can both be computed for any given $x$ using $\OO(t M^3)$ kernel evaluations, and describe all the 
implementation details of \kerlinexp.

Our main result regarding the performance of \kerlinexp for the two  different eigenvalue decay conditions is the following: 
\begin{theorem}\label{th_robust_kernel} 
Suppose that the kernel $\kappa$ satisfies Assumption~\ref{a:eigendecay} with the polynomial  eigenvalues decay rate $\mu_i  \le g i^{-c}$. 
Then, setting the parameters as  $M = T$, $\eta = \beta = T^{-\frac{1}{2}\pa{1+\frac{1}{c}}} \sqrt{\frac{(c-1)\ln T}{g}}$  the 
expected regret of \kerlinexp  satisfies
\begin{align*}
 		R_T = \OO\pa{ K T^{\frac{1}{2} \pa{1+\frac{1}{c}}}\sqrt{ (g/(c-1))\ln T}  }.
 \end{align*}
Furthermore, suppose that the kernel $\kappa$ satisfies Assumption~\ref{a:eigendecay} with the
 exponential decay rate $\mu_i  \le g e^{-c i}$. 
Then, setting the parameters as   
 $M = T$, $\eta = \beta = \sqrt{ \frac{ c\ln T}{g T} }$, the expected regret of \kerlinexp satisfies
\begin{align*}
 		R_T = \OO\pa{   K\sqrt{(g/c)  T \pa{\ln T}^3} }.
\end{align*}
\end{theorem}

\section{Analysis}\label{s:analysis}
In this section we provide the main arguments forming the proof of Theorem~\ref{th_robust_kernel}, relegating the 
proofs of some technical lemmas to Appendix~\ref{appx:proofs}. First, we introduce some important notations 
that will be useful throughout the proof. We first define the operator $\widehat{\Sigma}^{+}_{t,a} = I+ \sum_{k=1}^M 
C_{k}$ (with $C_k$ defined through the KGR subroutine for the $t,a$ pair in question), so that we can write the estimate of $f_{t,a}$ as 
\begin{equation}\label{eq:MGRestl}
	\widetilde{f}_{t,a} = \hSp_{t,a}  \varphi(X_t) \loss_t(X_t, A_t) \II{A_t= a}.
\end{equation}
Similarly, the exploration bonus $b_t(x,a)$ can be written using this notation as
\[
b_t(x,a) = \beta \norm{\varphi(x)}_{\Sigma_{t,a}^+}^2.
\]
Using this notation, we denote $\hloss_{t}(x,a) = \biprod{\varphi(x)}{\widetilde{f}_{t,a}} - b_t(x,a)$. 
When written in this form, it becomes readily apparent that our bonus is closely related to the adjustment proposed by \citet{BDHKRT08} for proving high-probability bounds in linear bandits (see also \citealp{ZL22}). That said, the purpose of our adjustment is quite different in that it mainly serves to remove a potentially harmful bias from the KGR estimators.
As for computing the estimates and bonuses defined above, note that the full functions $\widetilde{f}_{t,a}$ and $b_t$ are never computed by the algorithm, and are only evaluated at the 
contexts $X_{t+1},X_{t+2},\dots, X_T$ encountered in runtime. As explained in Appendix~\ref{app:computation}, each 
such evaluation has a cost of $\OO(tM^2)$. 

Our analysis will use ideas from \citet{NO20} and a number of new techniques that are necessary for dealing with the 
infinite-dimensional loss functions $f_{t,a}$. For the sake of analysis, we define $X_0$ as a sample from the context 
distribution $\mathcal{D}$ drawn independently from the history of interactions $\mathcal{F}_T$. We introduce the following notations:
\begin{itemize}[noitemsep]
	\item $\widetilde R_T =\EE{ \sum_{t=1}^T \sum_{a\in \A} (\pi_t(a|X_0) - \pi^*(a|X_0))\hloss_{t}(X_0,a)}$,
	\item $B^*_T= \EE{ \sum_{t=1}^T \sum_{a\in \A}  \pi^*(a|X_0) \pa{\hloss_{t}(X_0,a) - \loss_{t}(X_0,a)}} $,
	\item $B_T=  \EE{\sum_{t=1}^T \sum_{a\in \A} \pi_t(a|X_0)  \pa{\loss_{t}(X_0,a) - \hloss_{t}(X_0,a)} }$.
\end{itemize}
The first step in our proof is then to rewrite the regret as the sum of these three terms:
\begin{equation}\label{eq:regret_dec}
	R_T =  \widetilde R_T + B^*_T + B_T.
\end{equation}
The proof of this claim is a straightforward extension of the regret decomposition of Lemma~3 in \citet{NO20} and can be found in Appendix~\ref{appx:regret_dec}.

The terms in the decomposition can be 
interpreted as follows. First, $B_T^*$ is the \emph{overestimation bias} of the total loss of the comparator policy $\pi^*$, 
measuring the extent to which the expectation of the estimated loss of $\pi^*$ exceed the actual loss of the same 
policy. Similarly, $B_T$ is the \emph{underestimation bias} of the total loss incurred by the learner. As we will show, the 
overestimation bias can be uniformly upper bounded for all comparator policies thanks to the optimistic adjustment term $b_{t}(x,a)$ added to the loss function. Furthermore, we will show that the price of this adjustment is a term of the order $\EEt{b_t(X_t,A_t)} = \beta \EEt{\trace{\Sigma_{t,a}\Sigma_{t,a}^+}}$, which can be controlled in terms of the effective dimension of the kernel.

To provide an interpretation for the term $\widetilde{R}_T$, let's consider an auxiliary online learning problem where  
$x$  is fixed, there are $K$ actions,  and the losses are defined as $c_{t,a} = \hloss_{t}(x,a)$ for each $t, a$. We 
then execute a copy of FTRL with the log-barrier regularizer on this sequence of losses, resulting in the sequence of 
action distributions $\pi_t = \argmin_{p\in\Delta(\A)} \pa{\Psi(p) + \eta \sum_a \sum_{\tau=1}^t \hloss_{\tau}(x,a)}$. 
Thus, the regret in the auxiliary game against the comparator $\pi^*$ at $x$ can be expressed as
\begin{equation}\label{eq:x_regret}
	\wh{R}_T(x) = \sum_{t=1}^T\sum_{a} \bpa{\pi_t(a|x) - \pi^*(a|x)} \hloss_{t}(x,a).
\end{equation}
Now it is easy to notice that $\wt{R}_T$ can be expressed in terms of the regret in these auxiliary games as $\wt{R}_T 
= \EE{\wh{R}_T(X_0)}$. Our proof strategy will be to prove an almost-sure regret bound for the
auxiliary games defined at each $x$ and take expectation of the resulting bounds with respect to the law
of $X_0$, thus achieving a bound on the regret. 

Before we jump into the analysis of each term discussed above, we state a technical result that will be used repeatedly in nearly all proofs. The simple proof is provided in Appendix~\ref{appx:effective_dim_MGR}.
\begin{lemma}\label{lem:effective_dim_MGR} For all $t,a$ and $\varepsilon > 0$, we have
 \[
  \trace{\EEt{\Sigma_{t,a}^+ \Sigma_{t,a}} } = \trace{ (I - (I - \Sigma_{t,a})^M)} \le m(\varepsilon) + M \varepsilon.
 \]
\end{lemma}

\subsection{The bias of the loss estimator}
The most important ingredient in our analysis is establishing a bound on the bias of our loss estimators $\hloss_t$. 
The following lemma is our key tool that we use to this end.
\begin{lemma}\label{lem:trace_M}
	 For any $x\in \X$, for any $\beta >0, \gamma > 0, \lambda >0$, we have
	 \begin{equation}\label{trace_Mhat}
	 	\bigl|\EEtb{\biprod{\varphi(x)}{f_{t,a} - \wt{f}_{t,a}}}\bigr| \le \beta\EEt{\norm{\varphi(x)}^2_{ 
\widehat{\Sigma}_{t,a}^{+}}} + \frac{1}{\beta (M+1)}.
	 \end{equation}
\end{lemma}
The proof follows from a more or less straightforward calculation regarding the bias arising from the truncated 
geometric series we use to approximate the ``inverse'' of $\Sigma_{t,a}$. While the building blocks are standard, the 
result itself is new and and valuable in the sense that it gives a tighter control on the bias of the geometric 
resampling estimator than previous works (e.g.,~\citealp{NO20}). This tighter bound is enabled by our use of the 
log-barrier policy that allows us to set $M$ significantly larger than what the previous analysis of \citet{NO20} could 
have tolerated, which in turns enables meaningful control of the additional bias term $\frac{1}{\beta (M+1)}$ appearing 
in the above bound. We relegate the proof of this result to Appendix~\ref{appx:trace_M}.

We are now well-equipped to tackle the bias terms $B_T^*$ and $B_T$. We first show a bound on the overestimation bias:
\begin{lemma}\label{lem:overestimation}
The overestimation bias can be bounded as $
B_T^* \le   \frac{T}{\beta (M+1)}$.
\end{lemma}
\begin{proof}
 We appeal to Lemma~\ref{lem:trace_M} to show that
	\begin{align*}
	 	 	\EEtb{\hloss_t(x,a)} - \loss_t(x,a) 
	 	 	&= \iprod{\varphi(x)}{\EEtb{\wt{f}_{t,a}} - f_{t,a}} - \EEt{b_{t}(x,a)} \\
	 	 	&\le 
\beta\EEt{\norm{\varphi(x)}^2_{\widehat{\Sigma}_{t,a}^{+}}} + \frac{1}{\beta (M+1)} - \EEt{b_{t}(x,a)} = 
\frac{1}{\beta (M+1)},
	\end{align*}
where we recalled the definition of $b_t(x,a)$ in the last step. The claim then follows from averaging both sides with the joint distribution of $X_0$ and $A^*\sim\pi^*(X_0)$, and summing up for all $t$.
\end{proof}
Notice that without the optimistic adjustment $b_t(x,a)$, the overestimation bias would scale with 
$\EE{\sum_a \pi^*(a|X_0) \norm{\varphi(X_0)}^2_{ \widehat{\Sigma}_{t,a}^{+}} }$, which cannot be meaningfully bounded in 
general. 
The second lemma takes care of the underestimation bias, and also establishes the price of adding the exploration bonus 
$b_t(x,a)$ to the loss estimator.
\begin{lemma}\label{lem:underestimation}
 The underestimation bias can be bounded for any $\varepsilon>0$ as
\[
B_T \le 2(m(\varepsilon) + M\varepsilon) +  \frac{T}{\beta (M+1)}.
\]
\end{lemma}
\begin{proof}
 By applying Lemma~\ref{lem:trace_M}, we get that
	\begin{align*}
	 	 	&\loss_t(x,a) - \EEtb{\hloss_t(x,a)}
	 	 	= \iprod{\varphi(x)}{f_{t,a} - \EEtb{\wt{f}_{t,a}} } + \EEt{b_{t}(x,a)} \\
	 	 	&\qquad \qquad \le 
\beta\EEt{\norm{\varphi(x)}^2_{\widehat{\Sigma}_{t,a}^{+}}} + \frac{1}{\beta (M+1)} + \EEt{b_{t}(x,a)} 
= 2  \EEt{b_{t}(x,a)} + \frac{1}{\beta (M+1)},
	\end{align*}
where we recalled the definition of $b_t(x,a)$ in the last step. Taking expectations and averaging with respect to 
$\pi_t(\cdot|X_0)$ and the distribution of $X_0$, we get
\begin{align*}
	\EEt{  \pi_t(a|X_0) \norm{\varphi(X_0)}^2_{ \widehat{\Sigma}_{t,a}^{+}} } =  \EEt{\trace{\Sigma_{t,a} 
\widehat{\Sigma}_{t,a}^{+} }}  \le m(\varepsilon) + M\varepsilon,
\end{align*}
where the last step follows from an application of Lemma~\ref{lem:effective_dim_MGR}.
The proof os concluded by summing up for all $t$.
\end{proof}
In words, the effect of the optimistic bias is a factor of $2$ multiplying the term $\EEt{  \pi_t(a|X_0) 
\norm{\varphi(X_0)}^2_{ \widehat{\Sigma}_{t,a}^{+}} }$, which itself can be bounded effectively in terms of the 
effective dimension.

\subsection{Bounding the auxiliary regret}
The first major step in our proof is to bound the regret in the auxiliary games, which is done in the following standard 
lemma concerning the bound of FTRL with log-barrier regularization:
\begin{lemma}\label{l:log_barrier}
	Let $p_1, \dots, p_T \in \Delta(\A)$ be defined as
	\[
	p_t = \argmin_{p\in \Delta(\A)}\biggl\{ \eta \sum_a p_a \sum_{\tau \le t} c_{\tau,a}+\Psi(p)         \biggr\}, 
\forall t = 1, \dots, T,
	\]
	where $c_t \in \real^{\A}$ is an arbitrary loss vector and $\Psi(p) = \sum_{a\in\A} \ln \frac{1}{p_a}$. Then, for any $y\in \Delta(\A)$,
	\[  \sum_{t = 1}^T\sum_{a}\pa{p_{t,a} - y_a}c_{t,a} \le \frac{\Psi(y) - \Psi(p_1)}{\eta} + \eta 
\sum_{t=1}^{T}\sum_{a\in\A} p_{t,a}c^2_{t,a} .\]
\end{lemma}
The proof of this result is standard and can be found in a number of references---we point the reader to Lemma 3.1 from 
\cite{dai2023refined} for concreteness. 
Notably, the second term in this bound has the the same qualitative form as the standard bound for FTRL with negative 
entropy, with the key advantage that it does not require any assumptions regarding the range of losses $c_t$. 

Before we apply the above result to bounding $\wt{R}_T$, we state the following useful technical result regarding the 
second moment of the KGR estimator:
\begin{lemma}\label{l:variance}
	Suppose that $X_t$ satisfies $\norm{\varphi(X_t)} \le 1 $ for each $t$. Then for each $t$, 
the following inequality holds for any $\varepsilon > 0$:
	\label{quadratic} 
	$$\EEt{\sum_{a=1}^K    \pi_t(a|\tX)   \biprod{\varphi(\tX)}{\wt{f}_{t,a}}^2} \le  2K \pa{ 1  +  (m(\varepsilon ) + 
M\varepsilon)} .$$
\end{lemma}
The proof of this lemma follows from a rather tedious calculation that can be found in Appendix~\ref{appx:variance}.
With this result at hand, we are ready to state and prove the last remaining part of our regret bound.
\begin{lemma}\label{l:reg_term} For any positive $\eta, \beta, M, \varepsilon$,  \kerlinexp guarantees
    \begin{align*}
    \widetilde R_{T} \le \frac{K \ln T}{\eta} +  2 + 2\beta(M+1) + \frac{2}{\beta (M+1)} + 2 \eta KT \pa{ 2
 +  \pa{m(\varepsilon ) + M\varepsilon}} \cdot\pa{2+\beta^2 M}.
	\end{align*}
\end{lemma}
\begin{proof}
Let us fix $x\in\X$ and apply Lemma~\ref{l:log_barrier} to obtain the following:
	\begin{equation}
	\begin{split}
	\label{eq:regret_decomp}
		&\sum_{t = 1}^T \sum_{a\in\A} (\pi_t(a|x) - \pi^*(a|x))\pa{\hat{\ell}_k(a,x) } \le 
\frac{\Psi(\tilde\pi^*(\cdot|x)) - \Psi(\pi_1(\cdot|x))}{\eta} \\
		& \qquad +\sum_{t=1}^T \sum_{a\in\A} ( \tilde\pi^*(a|x)  -   \pi^*(a|x) )\pa{ \hloss_t(  x,a) }  + \eta 
\sum_{t=1}^T \sum_{a\in\A} \pi_t(a|x)\pa{ \hloss_t(  x,a) }^2.
		\end{split}
\end{equation}
	By picking $\tilde\pi^*(a|x) = \pa{1 - \frac{K}{T}}\pi^*(a|x) + \frac{1}{T}$, the first term is bounded by 
	\begin{align*}
		 \frac{\Psi(\tilde\pi^*(\cdot|x)) - \Psi(\pi_1(\cdot|x))}{\eta}  \le \frac{K \ln T}{\eta}.
	\end{align*}
	To proceed, we appeal to Lemma~\ref{lem:trace_M} to show that
	\[
	 	 	\bigl|\EEtb{\hloss_t(x,a)} - \loss_t(x,a)\bigr| \le 2\beta\EEt{\norm{\varphi(x)}^2_{ 
\widehat{\Sigma}_{t,a}^{+}}} + \frac{1}{\beta (M+1)} = 2\EEt{b_{t}(x,a)} + \frac{1}{\beta (M+1)},
	\]
	and also observe that the exploration bonus can be bounded as
	\[
	 b_{t}(x,a) = \beta \norm{\varphi(x)}^2_{\hat{\Sigma}_{t,a}^+} \le \beta \norm{\wh{\Sigma}_{t,a}^+}_2  \le 
\beta (M+1).
	\]
	Altogether, these observations can be used to simply bound the second term on the right-hand side of Equation~\eqref{eq:regret_decomp} as
	\begin{align*}
		 &\EE{\sum_{t=1}^T \sum_{a\in\A} ( \tilde\pi^*(a|x)  -   \pi^*(a|x) )\hloss_t(x,a)  } 
 \le
 2\pa{1 + \beta(M+1) + \frac{1}{\beta (M+1)}}.
	\end{align*}
It remains to bound the last term on the right-hand side of Equation~\eqref{eq:regret_decomp}, which we start by 
writing
	\begin{align*}
		&\eta \sum_{t=1}^T \sum_{a\in\A} \pi_t(a|x)\pa{ \hloss_t(  x,a) }^2\le 2\eta  \sum_{t=1}^T \sum_{a\in\A} \pi_t(a|x)\pa{\pa{\biprod{\varphi(x)}{\wt{f}_{t,a}}}^2 +\pa{ b_t(  x,a)  }^2}.
	\end{align*}
	We will bound these terms on expectation with respect to the random context $X_0$.
The first term in the resulting expression can be upper bounded by using Lemma~\ref{l:variance}:
			$$\EEt{\sum_{t=1}^T\sum_{a=1}^A   \pi_t(a|\tX)   \biprod{\varphi(\tX)}{\wt{f}_{t,a}}^2} \le  2K \pa{ 1  +  
(m(\varepsilon ) + 
M\varepsilon)}.$$
	Moving on to the second term, we have
	\begin{align*}
		&\EEt{ \sum_{t=1}^T \sum_{a\in\A} \pi_t(a|\tX)\pa{ b_t(  \tX,a)  }^2} 
  \le  \beta^2 M  \EEt{ \sum_{t=1}^T \sum_{a\in\A} \pi_t(a|\tX) \norm{\varphi(X_0)}_{\hSigma^+_{t,a}}^2}\\
		&\qquad\qquad=  \beta^2 M\EEt{ \sum_{t=1}^T \sum_{a\in\A}  \trace{\Sigma_{t,a} \hSigma^+_{t,a} }}
		\le \beta^2 M  \pa{m(\varepsilon) + M\varepsilon} KT,
	\end{align*}
	where the last inequality uses Lemma~\ref{lem:effective_dim_MGR}. Collecting all terms together, we get
	\begin{align*}
	\widetilde R_{T} \le \frac{K \ln T}{\eta} +  2 + 2\beta(M+1) + \frac{2}{\beta (M+1)} + 2 \eta KT \pa{ 2
 +  \pa{m(\varepsilon ) + M\varepsilon}} \cdot\pa{2+\beta^2 M},
\end{align*}
thus concluding the proof.
\end{proof}	

\subsection{The proof of Theorem~\ref{th_robust_kernel}}
The proof now follows from putting together the results of Lemmas~\ref{lem:overestimation}, \ref{lem:underestimation}, 
and~\ref{l:reg_term}, yielding
\begin{align*}
    R_T & = \wt{R}_T + B_T^* + B_T \\
    &\le \frac{K \ln T}{\eta} +  2 + 2\beta(M+1) + \frac{2}{\beta (M+1)} + 2 \eta KT \pa{ 2
	 +  \pa{m(\varepsilon ) + M\varepsilon}} \cdot\pa{2+\beta^2 M} \\
	 &+ \frac{2T}{\beta (M+1) } + 2\beta K (m(\varepsilon)+M\varepsilon)T.
\end{align*}

It remains to derive the concrete rates claimed in the theorem for the two separate eigendecay regimes considered 
 therein. First, consider the 
 polynomial  decay rate $\mu_i  \le gi^{-c}$ and recall from Section~\ref{sec:kernels} that we have $m(\varepsilon) = \OO\pa{((c-1)\varepsilon/g)^{1/(c-1)}}$ in this case. Thus, we can set $\varepsilon = \frac{g}{c-1}T^{\frac{1-c}{c}}$ which yields $m=\OO\pa{T^{1/c}}$.  Taking  $M = T$, $\eta = \beta = T^{-\frac{1}{2}\pa{1+\frac{1}{c}}} \sqrt{\frac{(c-1)\ln T}{g}}$ and plugging into the bound above, the expected
 	regret of \kerlinexp  can be seen to satisfy
 	\begin{align*}
 		R_T = \OO\pa{ K \sqrt{ (g/(c-1))\ln(T)}  T^{\frac{1}{2} \pa{1+\frac{1}{c}}}},
 	\end{align*}
 	proving the first claim. 
 
 As for  the exponential decay $\mu_i  \le g e^{-c i}$, recall from Section~\ref{sec:kernels} that $m(\varepsilon) = \OO\pa{\frac{\ln\pa{g/(c\varepsilon)}}{c}}$, so that we can set $\varepsilon=\frac{g}{cT}$ to get $m(\varepsilon) = \frac{\ln T}{c}$. Letting  $M = T$, $\eta = \beta = \sqrt{ \frac{ c\ln T}{g T} }$, and substituting these values into the previously derived bound, we can observe that  
 	\begin{align*}
 		R_T = \OO\pa{   K\sqrt{(g/c)  T \ln(T) } \ln(T) },
 	\end{align*}
	which proves the second  claim. 
\qed


\section{Discussion}\label{s:discussion}

    We now turn to discussing our results in some more detail, focusing on comparison with related work and the possibility to improve certain aspects of our algorithm and its theoretical guarantees. 

    The first question one may ask is if our results match the best achievable regret bounds in this context. While we cannot provide a fully affirmative answer to this question, there are definitely reasons to believe that at least the dependence of our bounds on $T$ is optimal. In the special cases of Mat\'ern and Gaussian kernels, our upper bounds match the lower bounds proved by \citet{SBC17} and also the lower bounds of \citet{CPB19} that were proved for more general kernels but in a slightly different setting. In the general case, our bounds can also be shown to match the best known rates for the stochastic version of our problem claimed by \citet{VKMFC13}---see the discussion in Appendix~D of \citet{ZBDMMG22} that relates the various notions of ``effective dimension'' used in these works. A comparison with these results is made possible by noticing that $\trace{I - (I-\Sigma_{t,a})^M}$ can be also seen as an effective dimension that closely matches the other dimensions proposed in the previously mentioned papers \citep{YRC07,RWY14}. In light of these observations, we conjecture that our bounds are optimal in terms of $T$ under the set of assumptions we make.

    One remarkable downside of our bounds is their linear scaling with the number of actions $K$. This obviously suboptimal scaling is due to the use of the log-barrier regularizer in our algorithm. We conjecture that this factor can be improved by a more sophisticated algorithm design. One potential idea that we believe could work would be to adapt the very recently proposed ``magnitude-reduced'' loss estimators of \citet{dai2023refined} in tandem with a standard entropy regularizer, but we can see many potential failure modes for this approach and as such we leave its exploration for future work.



Finally, let us comment on the computational complexity of our method. In Appendix~\ref{app:computation}, we show that the computation of $\wh{L}_{t-1,a}$ for all actions  takes $\OO(K(t-1) M^3)$ steps, which makes for a total computational complexity of $\OO(KT^5)$ over $T$ rounds due to our choice of $M=T$. While polynomial in $T$, this rate is obviously not the most practical that one can wish for, and thus it is a natural question to ask if a faster method can be devised without compromising the regret bounds. A potential idea to consider is to use sketching methods such as the ones used by \citet{CCLVR19,CCLVR20} or \citet{ZBDMMG22} to reduce the computational burden. We note that it is not obvious at all if such methods can achieve the desired goal, as none of these sketching-based methods are able to attain the near-optimal rates of inefficient algorithms like that of \citet{VKMFC13}.
	


\bibliography{references.bib}

\begin{thebibliography}{20}
\providecommand{\natexlab}[1]{#1}
\providecommand{\url}[1]{\texttt{#1}}
\expandafter\ifx\csname urlstyle\endcsname\relax
  \providecommand{\doi}[1]{doi: #1}\else
  \providecommand{\doi}{doi: \begingroup \urlstyle{rm}\Url}\fi

\bibitem[Abbasi-Yadkori et~al.(2011)Abbasi-Yadkori, P{\'a}l, and
  Szepesv{\'a}ri]{abbasi2011improved}
Yasin Abbasi-Yadkori, D{\'a}vid P{\'a}l, and Csaba Szepesv{\'a}ri.
\newblock Improved algorithms for linear stochastic bandits.
\newblock \emph{Advances in neural information processing systems}, 24, 2011.

\bibitem[Auer et~al.(2002)Auer, Cesa-Bianchi, Freund, and
  Schapire]{auer2002nonstochastic}
Peter Auer, Nicolo Cesa-Bianchi, Yoav Freund, and Robert~E Schapire.
\newblock The nonstochastic multiarmed bandit problem.
\newblock \emph{SIAM journal on computing}, 32\penalty0 (1):\penalty0 48--77,
  2002.

\bibitem[Beygelzimer et~al.(2011)Beygelzimer, Langford, Li, Reyzin, and
  Schapire]{beygelzimer2011contextual}
Alina Beygelzimer, John Langford, Lihong Li, Lev Reyzin, and Robert Schapire.
\newblock Contextual bandit algorithms with supervised learning guarantees.
\newblock In \emph{Proceedings of the Fourteenth International Conference on
  Artificial Intelligence and Statistics}, pages 19--26. JMLR Workshop and
  Conference Proceedings, 2011.

\bibitem[Chatterji et~al.(2019)Chatterji, Pacchiano, and Bartlett]{CPB19}
Niladri Chatterji, Aldo Pacchiano, and Peter Bartlett.
\newblock Online learning with kernel losses.
\newblock In Kamalika Chaudhuri and Ruslan Salakhutdinov, editors,
  \emph{Proceedings of the 36th International Conference on Machine Learning},
  volume~97 of \emph{Proceedings of Machine Learning Research}, pages 971--980.
  PMLR, 09--15 Jun 2019.
\newblock URL \url{https://proceedings.mlr.press/v97/chatterji19a.html}.

\bibitem[Chowdhury and Gopalan(2017)]{chowdhury2017kernelized}
Sayak~Ray Chowdhury and Aditya Gopalan.
\newblock On kernelized multi-armed bandits.
\newblock In \emph{International Conference on Machine Learning}, pages
  844--853. PMLR, 2017.

\bibitem[Christmann and Steinwart(2008)]{Christmann2008}
Andreas Christmann and Ingo Steinwart.
\newblock \emph{Support Vector Machines}.
\newblock Springer New York, NY, 2008.

\bibitem[Chu et~al.(2011)Chu, Li, Reyzin, and Schapire]{chu2011contextual}
Wei Chu, Lihong Li, Lev Reyzin, and Robert Schapire.
\newblock Contextual bandits with linear payoff functions.
\newblock In \emph{Proceedings of the Fourteenth International Conference on
  Artificial Intelligence and Statistics}, pages 208--214. JMLR Workshop and
  Conference Proceedings, 2011.

\bibitem[Dai et~al.(2023)Dai, Luo, Wei, and Zimmert]{dai2023refined}
Yan Dai, Haipeng Luo, Chen-Yu Wei, and Julian Zimmert.
\newblock Refined regret for adversarial mdps with linear function
  approximation.
\newblock \emph{arXiv preprint arXiv:2301.12942}, 2023.

\bibitem[Foster et~al.(2016)Foster, Li, Lykouris, Sridharan, and
  Tardos]{foster2016learning}
Dylan~J Foster, Zhiyuan Li, Thodoris Lykouris, Karthik Sridharan, and Eva
  Tardos.
\newblock Learning in games: Robustness of fast convergence.
\newblock \emph{Advances in Neural Information Processing Systems}, 29, 2016.

\bibitem[Foster et~al.(2020)Foster, Gentile, Mohri, and
  Zimmert]{foster2020adapting}
Dylan~J Foster, Claudio Gentile, Mehryar Mohri, and Julian Zimmert.
\newblock Adapting to misspecification in contextual bandits.
\newblock \emph{Advances in Neural Information Processing Systems},
  33:\penalty0 11478--11489, 2020.

\bibitem[Li et~al.(2019)Li, Wang, and Zhou]{li2019nearly}
Yingkai Li, Yining Wang, and Yuan Zhou.
\newblock Nearly minimax-optimal regret for linearly parameterized bandits.
\newblock In \emph{Conference on Learning Theory}, pages 2173--2174. PMLR,
  2019.

\bibitem[Liu et~al.(2023)Liu, Wei, and Zimmert]{liu2023bypassing}
Haolin Liu, Chen-Yu Wei, and Julian Zimmert.
\newblock Bypassing the simulator: Near-optimal adversarial linear contextual
  bandits.
\newblock \emph{arXiv preprint arXiv:2309.00814}, 2023.

\bibitem[Mercer(1909)]{Mercer1909}
J.~Mercer.
\newblock Functions of positive and negative type, and their connection with
  the theory of integral equations.
\newblock \emph{Philosophical Transactions of the Royal Society of London.
  Series A, Containing Papers of a Mathematical or Physical Character},
  209:\penalty0 415--446, 1909.
\newblock ISSN 02643952.
\newblock URL \url{http://www.jstor.org/stable/91043}.

\bibitem[Neu and Olkhovskaya(2020)]{NO20}
Gergely Neu and Julia Olkhovskaya.
\newblock Efficient and robust algorithms for adversarial linear contextual
  bandits.
\newblock In \emph{Conference on Learning Theory}, pages 3049--3068. PMLR,
  2020.

\bibitem[Rakhlin and Sridharan(2016)]{rakhlin2016bistro}
Alexander Rakhlin and Karthik Sridharan.
\newblock Bistro: An efficient relaxation-based method for contextual bandits.
\newblock In \emph{International Conference on Machine Learning}, pages
  1977--1985. PMLR, 2016.

\bibitem[Seeger et~al.(2008)Seeger, Kakade, and Foster]{SMKF08}
Matthias~W. Seeger, Sham~M. Kakade, and Dean~P. Foster.
\newblock Information consistency of nonparametric gaussian process methods.
\newblock \emph{IEEE Transactions on Information Theory}, 54\penalty0
  (5):\penalty0 2376--2382, 2008.
\newblock \doi{10.1109/TIT.2007.915707}.

\bibitem[Srinivas et~al.(2009)Srinivas, Krause, Kakade, and
  Seeger]{srinivas2009gaussian}
Niranjan Srinivas, Andreas Krause, Sham~M Kakade, and Matthias Seeger.
\newblock Gaussian process optimization in the bandit setting: No regret and
  experimental design.
\newblock \emph{arXiv preprint arXiv:0912.3995}, 2009.

\bibitem[Syrgkanis et~al.(2016)Syrgkanis, Krishnamurthy, and
  Schapire]{syrgkanis2016efficient}
Vasilis Syrgkanis, Akshay Krishnamurthy, and Robert Schapire.
\newblock Efficient algorithms for adversarial contextual learning.
\newblock In \emph{International Conference on Machine Learning}, pages
  2159--2168. PMLR, 2016.

\bibitem[Tewari and Murphy(2017)]{Tewari2017}
Ambuj Tewari and Susan~A. Murphy.
\newblock From ads to interventions: Contextual bandits in mobile health.
\newblock In \emph{Mobile Health - Sensors, Analytic Methods, and
  Applications}, 2017.
\newblock URL \url{https://api.semanticscholar.org/CorpusID:18778220}.

\bibitem[Yang et~al.(2020)Yang, Jin, Wang, Wang, and Jordan]{yang2020function}
Zhuoran Yang, Chi Jin, Zhaoran Wang, Mengdi Wang, and Michael~I Jordan.
\newblock On function approximation in reinforcement learning: Optimism in the
  face of large state spaces.
\newblock \emph{Advances in Neural Information Processing Systems}, 2020, 2020.

\end{thebibliography}


\begin{thebibliography}{38}
\providecommand{\natexlab}[1]{#1}
\providecommand{\url}[1]{\texttt{#1}}
\expandafter\ifx\csname urlstyle\endcsname\relax
  \providecommand{\doi}[1]{doi: #1}\else
  \providecommand{\doi}{doi: \begingroup \urlstyle{rm}\Url}\fi

\bibitem[Abbasi-Yadkori et~al.(2011)Abbasi-Yadkori, P{\'a}l, and
  Szepesv{\'a}ri]{abbasi2011improved}
Yasin Abbasi-Yadkori, D{\'a}vid P{\'a}l, and Csaba Szepesv{\'a}ri.
\newblock Improved algorithms for linear stochastic bandits.
\newblock \emph{Advances in neural information processing systems}, 24, 2011.

\bibitem[Agarwal et~al.(2017)Agarwal, Luo, Neyshabur, and Schapire]{ALNS17}
Alekh Agarwal, Haipeng Luo, Behnam Neyshabur, and Robert~E Schapire.
\newblock Corralling a band of bandit algorithms.
\newblock In \emph{Conference on Learning Theory}, pages 12--38, 2017.

\bibitem[Auer et~al.(2002)Auer, Cesa-Bianchi, Freund, and
  Schapire]{auer2002nonstochastic}
Peter Auer, Nicolo Cesa-Bianchi, Yoav Freund, and Robert~E Schapire.
\newblock The nonstochastic multiarmed bandit problem.
\newblock \emph{SIAM journal on computing}, 32\penalty0 (1):\penalty0 48--77,
  2002.

\bibitem[Awasthi et~al.(2015)Awasthi, Charikar, Lai, and Risteski]{ACLR15}
Pranjal Awasthi, Moses Charikar, Kevin~A. Lai, and Andrej Risteski.
\newblock Label optimal regret bounds for online local learning.
\newblock In \emph{Proceedings of The 28th Conference on Learning Theory
  (COLT)}, pages 150--166, 2015.

\bibitem[Bartlett et~al.(2008)Bartlett, Dani, Hayes, Kakade, Rakhlin, and
  Tewari]{BDHKRT08}
Peter Bartlett, Varsha Dani, Thomas Hayes, Sham Kakade, Alexander Rakhlin, and
  Ambuj Tewari.
\newblock High-probability regret bounds for bandit online linear optimization.
\newblock In \emph{Proceedings of the 21st Annual Conference on Learning
  Theory-COLT 2008}, pages 335--342, 2008.

\bibitem[Beygelzimer et~al.(2011)Beygelzimer, Langford, Li, Reyzin, and
  Schapire]{beygelzimer2011contextual}
Alina Beygelzimer, John Langford, Lihong Li, Lev Reyzin, and Robert Schapire.
\newblock Contextual bandit algorithms with supervised learning guarantees.
\newblock In \emph{Proceedings of the Fourteenth International Conference on
  Artificial Intelligence and Statistics}, pages 19--26. JMLR Workshop and
  Conference Proceedings, 2011.

\bibitem[Bubeck et~al.(2018)Bubeck, Cohen, and Li]{BCL18}
S{\'e}bastien Bubeck, Michael Cohen, and Yuanzhi Li.
\newblock Sparsity, variance and curvature in multi-armed bandits.
\newblock In \emph{Algorithmic Learning Theory}, pages 111--127, 2018.

\bibitem[Calandriello et~al.(2019)Calandriello, Carratino, Lazaric, Valko, and
  Rosasco]{CCLVR19}
Daniele Calandriello, Luigi Carratino, Alessandro Lazaric, Michal Valko, and
  Lorenzo Rosasco.
\newblock Gaussian process optimization with adaptive sketching: Scalable and
  no regret.
\newblock In \emph{Conference on Learning Theory}, pages 533--557, 2019.

\bibitem[Calandriello et~al.(2020)Calandriello, Carratino, Lazaric, Valko, and
  Rosasco]{CCLVR20}
Daniele Calandriello, Luigi Carratino, Alessandro Lazaric, Michal Valko, and
  Lorenzo Rosasco.
\newblock Near-linear time gaussian process optimization with adaptive batching
  and resparsification.
\newblock In \emph{International Conference on Machine Learning}, pages
  1295--1305, 2020.

\bibitem[Chatterji et~al.(2019)Chatterji, Pacchiano, and Bartlett]{CPB19}
Niladri Chatterji, Aldo Pacchiano, and Peter Bartlett.
\newblock Online learning with kernel losses.
\newblock In Kamalika Chaudhuri and Ruslan Salakhutdinov, editors,
  \emph{Proceedings of the 36th International Conference on Machine Learning},
  volume~97 of \emph{Proceedings of Machine Learning Research}, pages 971--980.
  PMLR, 09--15 Jun 2019.
\newblock URL \url{https://proceedings.mlr.press/v97/chatterji19a.html}.

\bibitem[Chowdhury and Gopalan(2017)]{chowdhury2017kernelized}
Sayak~Ray Chowdhury and Aditya Gopalan.
\newblock On kernelized multi-armed bandits.
\newblock In \emph{International Conference on Machine Learning}, pages
  844--853. PMLR, 2017.

\bibitem[Christiano(2016)]{Chr16}
Paul Christiano.
\newblock Provably manipulation-resistant reputation systems.
\newblock In \emph{Proceedings of the 29th Annual Conference on Learning Theory
  (COLT)}, pages 670--697, 2016.

\bibitem[Christmann and Steinwart(2008)]{Christmann2008}
Andreas Christmann and Ingo Steinwart.
\newblock \emph{Support Vector Machines}.
\newblock Springer New York, NY, 2008.

\bibitem[Chu et~al.(2011)Chu, Li, Reyzin, and Schapire]{chu2011contextual}
Wei Chu, Lihong Li, Lev Reyzin, and Robert Schapire.
\newblock Contextual bandits with linear payoff functions.
\newblock In \emph{Proceedings of the Fourteenth International Conference on
  Artificial Intelligence and Statistics}, pages 208--214. JMLR Workshop and
  Conference Proceedings, 2011.

\bibitem[Dai et~al.(2023)Dai, Luo, Wei, and Zimmert]{dai2023refined}
Yan Dai, Haipeng Luo, Chen-Yu Wei, and Julian Zimmert.
\newblock Refined regret for adversarial mdps with linear function
  approximation.
\newblock \emph{arXiv preprint arXiv:2301.12942}, 2023.

\bibitem[Davis et~al.(2007)Davis, Kulis, Jain, Sra, and Dhillon]{DKJSD07}
Jason~V Davis, Brian Kulis, Prateek Jain, Suvrit Sra, and Inderjit~S Dhillon.
\newblock Information-theoretic metric learning.
\newblock In \emph{Proceedings of the 24th international conference on Machine
  learning}, pages 209--216. ACM, 2007.

\bibitem[Foster et~al.(2016)Foster, Li, Lykouris, Sridharan, and
  Tardos]{foster2016learning}
Dylan~J Foster, Zhiyuan Li, Thodoris Lykouris, Karthik Sridharan, and Eva
  Tardos.
\newblock Learning in games: Robustness of fast convergence.
\newblock \emph{Advances in Neural Information Processing Systems}, 29, 2016.

\bibitem[Foster et~al.(2020)Foster, Gentile, Mohri, and
  Zimmert]{foster2020adapting}
Dylan~J Foster, Claudio Gentile, Mehryar Mohri, and Julian Zimmert.
\newblock Adapting to misspecification in contextual bandits.
\newblock \emph{Advances in Neural Information Processing Systems},
  33:\penalty0 11478--11489, 2020.

\bibitem[Jain et~al.(2009)Jain, Kulis, Dhillon, and Grauman]{JKDG09}
Prateek Jain, Brian Kulis, Inderjit~S Dhillon, and Kristen Grauman.
\newblock Online metric learning and fast similarity search.
\newblock In \emph{Advances in neural information processing systems}, pages
  761--768, 2009.

\bibitem[Kulis and Bartlett(2010)]{KB10}
Brian Kulis and Peter~L Bartlett.
\newblock Implicit online learning.
\newblock In \emph{Proceedings of the 27th International Conference on Machine
  Learning (ICML-10)}, pages 575--582, 2010.

\bibitem[Li et~al.(2019)Li, Wang, and Zhou]{li2019nearly}
Yingkai Li, Yining Wang, and Yuan Zhou.
\newblock Nearly minimax-optimal regret for linearly parameterized bandits.
\newblock In \emph{Conference on Learning Theory}, pages 2173--2174. PMLR,
  2019.

\bibitem[Liu et~al.(2023)Liu, Wei, and Zimmert]{liu2023bypassing}
Haolin Liu, Chen-Yu Wei, and Julian Zimmert.
\newblock Bypassing the simulator: Near-optimal adversarial linear contextual
  bandits.
\newblock \emph{arXiv preprint arXiv:2309.00814}, 2023.

\bibitem[Luo et~al.(2018)Luo, Wei, and Zheng]{LWZ18}
Haipeng Luo, Chen-Yu Wei, and Kai Zheng.
\newblock Efficient online portfolio with logarithmic regret.
\newblock \emph{Advances in neural information processing systems}, 31, 2018.

\bibitem[Mercer(1909)]{Mercer1909}
J.~Mercer.
\newblock Functions of positive and negative type, and their connection with
  the theory of integral equations.
\newblock \emph{Philosophical Transactions of the Royal Society of London.
  Series A, Containing Papers of a Mathematical or Physical Character},
  209:\penalty0 415--446, 1909.
\newblock ISSN 02643952.
\newblock URL \url{http://www.jstor.org/stable/91043}.

\bibitem[Neu and Olkhovskaya(2020)]{NO20}
Gergely Neu and Julia Olkhovskaya.
\newblock Efficient and robust algorithms for adversarial linear contextual
  bandits.
\newblock In \emph{Conference on Learning Theory}, pages 3049--3068. PMLR,
  2020.

\bibitem[Raskutti et~al.(2014)Raskutti, Wainwright, and Yu]{RWY14}
Garvesh Raskutti, Martin~J Wainwright, and Bin Yu.
\newblock Early stopping and non-parametric regression: an optimal
  data-dependent stopping rule.
\newblock \emph{The Journal of Machine Learning Research}, 15\penalty0
  (1):\penalty0 335--366, 2014.

\bibitem[Rasmussen and Williams(2006)]{RW06}
Carl~Edward Rasmussen and Christopher~KI Williams.
\newblock \emph{Gaussian processes for machine learning}.
\newblock MIT press Cambridge, MA, 2006.

\bibitem[Scarlett et~al.(2017)Scarlett, Bogunovic, and Cevher]{SBC17}
Jonathan Scarlett, Ilija Bogunovic, and Volkan Cevher.
\newblock Lower bounds on regret for noisy gaussian process bandit
  optimization.
\newblock In \emph{Conference on Learning Theory}, pages 1723--1742, 2017.

\bibitem[Sch{\"o}lkopf and Smola(2002)]{SS02}
Bernhard Sch{\"o}lkopf and Alexander~J Smola.
\newblock \emph{Learning with kernels: support vector machines, regularization,
  optimization, and beyond}.
\newblock MIT press, 2002.

\bibitem[Seeger et~al.(2008)Seeger, Kakade, and Foster]{SMKF08}
Matthias~W. Seeger, Sham~M. Kakade, and Dean~P. Foster.
\newblock Information consistency of nonparametric gaussian process methods.
\newblock \emph{IEEE Transactions on Information Theory}, 54\penalty0
  (5):\penalty0 2376--2382, 2008.
\newblock \doi{10.1109/TIT.2007.915707}.

\bibitem[Srinivas et~al.(2009)Srinivas, Krause, Kakade, and
  Seeger]{srinivas2009gaussian}
Niranjan Srinivas, Andreas Krause, Sham~M Kakade, and Matthias Seeger.
\newblock Gaussian process optimization in the bandit setting: No regret and
  experimental design.
\newblock \emph{arXiv preprint arXiv:0912.3995}, 2009.

\bibitem[Tewari and Murphy(2017)]{Tewari2017}
Ambuj Tewari and Susan~A. Murphy.
\newblock From ads to interventions: Contextual bandits in mobile health.
\newblock In \emph{Mobile Health - Sensors, Analytic Methods, and
  Applications}, 2017.
\newblock URL \url{https://api.semanticscholar.org/CorpusID:18778220}.

\bibitem[Valko et~al.(2013)Valko, Korda, Munos, Flaounas, and
  Cristianini]{VKMFC13}
Michal Valko, Nathan Korda, R{\'e}mi Munos, Ilias Flaounas, and Nello
  Cristianini.
\newblock Finite-time analysis of kernelised contextual bandits.
\newblock In \emph{Uncertainty in Artificial Intelligence}, 2013.

\bibitem[Wei and Luo(2018)]{WL18}
Chen-Yu Wei and Haipeng Luo.
\newblock More adaptive algorithms for adversarial bandits.
\newblock In \emph{Conference On Learning Theory}, pages 1263--1291. PMLR,
  2018.

\bibitem[Yang et~al.(2020)Yang, Jin, Wang, Wang, and Jordan]{yang2020function}
Zhuoran Yang, Chi Jin, Zhaoran Wang, Mengdi Wang, and Michael~I Jordan.
\newblock On function approximation in reinforcement learning: Optimism in the
  face of large state spaces.
\newblock \emph{Advances in Neural Information Processing Systems}, 2020, 2020.

\bibitem[Yao et~al.(2007)Yao, Rosasco, and Caponnetto]{YRC07}
Yuan Yao, Lorenzo Rosasco, and Andrea Caponnetto.
\newblock On early stopping in gradient descent learning.
\newblock \emph{Constructive Approximation}, 26:\penalty0 289--315, 2007.

\bibitem[Zenati et~al.(2022)Zenati, Bietti, Diemert, Mairal, Martin, and
  Gaillard]{ZBDMMG22}
Houssam Zenati, Alberto Bietti, Eustache Diemert, Julien Mairal, Matthieu
  Martin, and Pierre Gaillard.
\newblock Efficient kernelized ucb for contextual bandits.
\newblock In \emph{International Conference on Artificial Intelligence and
  Statistics}, pages 5689--5720, 2022.

\bibitem[Zimmert and Lattimore(2022)]{ZL22}
Julian Zimmert and Tor Lattimore.
\newblock Return of the bias: Almost minimax optimal high probability bounds
  for adversarial linear bandits.
\newblock In \emph{Proceedings of Thirty Fifth Conference on Learning Theory},
  pages 3285--3312, 2022.

\end{thebibliography}

\appendix
\section{Mercer's theorem}
\label{appx:mercer}


Mercer's theorem \citep{Mercer1909} provides a representation of a positive-definite kernel $\kappa$ in terms of an infinite dimensional feature map (see, e.g. \cite{Christmann2008}, Theorem 4.49). Let $\mathcal{X}$ be a compact metric space and $\nu$ be a finite Borel measure on $\mathcal{X}$ (we consider Lebesgue measure in a Euclidean space). 
Let $L^2_\nu(\mathcal{X})$ be the set of square-integrable functions on $\mathcal{X}$ with respect to $\nu$. We further say that the kernel $\kappa$ square-integrable if
\begin{equation}
\int_{\mathcal{X}} \int_{\mathcal{X}} \kappa(x, x')^2 \,d \nu(x) d \nu(x')<\infty.
\end{equation}

\begin{theorem}
(Mercer's Theorem) Let $\mathcal{X}$ be a compact metric space and $\nu$ be a finite Borel measure on $\mathcal{X}$. Let $\kappa$ be a continuous and square-integrable kernel, inducing an integral operator $T_k:L^2_\nu(\mathcal{X})\rightarrow L^2_\nu(\mathcal{X})$ defined by
\begin{equation}
\left(T_\kappa f\right)(\cdot)=\int_{\mathcal{X}} \kappa(\cdot, x') f(x') \,d \nu(x')\,,
\end{equation}
where $f\in L^2_\nu(\mathcal{X})$. Then, there exists a sequence of eigenvalue-eigenfunction pairs $\left\{(\mu_i, \psi_i)\right\}_{i=1}^{\infty}$ such that $\mu_i >0$, and $T_\kappa \psi_i=\mu_i \psi_i$, for $m \geq 1$. Moreover, the kernel function can be represented as
\begin{equation}
\kappa\left(x, x'\right)=\sum_{i=1}^{\infty} \mu_i \psi_i(x) \psi_i\left(x'\right),
\end{equation}
where the convergence of the series holds uniformly on $\mathcal{X} \times \mathcal{X}$.
\end{theorem}

Additionally, the Mercer representation theorem (see, e.g., \cite{Christmann2008}, Theorem 4.51) states that the RKHS induced by $\kappa$ can consequently be represented in terms of $\{(\mu_i,\psi_i)\}_{i=1}^\infty$.

\begin{theorem}(Mercer Representation Theorem) Let $\left\{\left(\mu_i,\psi_i\right)\right\}_{i=1}^{\infty}$ be the Mercer eigenvalue-eigenfunction pairs. Then, the RKHS associated with $\kappa$ is given by
\begin{equation}
\mathcal{H}_\kappa=\left\{f(\cdot)=\sum_{i=1}^{\infty} w_i \lambda_i^{\frac{1}{2}} \psi_i(\cdot): w_i \in \mathbb{R},\|f\|_{\mathcal{H}_\kappa}^2:=\sum_{i=1}^{\infty} w_i^2<\infty\right\}
\end{equation}
\end{theorem}
In particular, the Mercer representation theorem indicates that the scaled eigenfunctions $\{\sqrt{\lambda_i}\psi_i\}_{i=1}^\infty$ form an orthonormal basis for $\Hc_\kappa$.

\section{Omitted proofs}
\allowdisplaybreaks
\label{appx:proofs}

\subsection{Proof of the regret decomposition (\ref{eq:regret_dec})}\label{appx:regret_dec}
Let us rewrite the estimate $\hloss_{t,a}(x) = \biprod{\varphi(x)}{ \wt f^*_{t,a}} +  \iprod{\varphi(x)}{\delta_{t,a}}  + b_{t,a}(x) $, where $\wt f^*_{t,a}$ and $\delta_{t,a}$ are such that $\EEt{\wt f^*_{t,a}} = f_{t,a}$ and $\widetilde{f}_{t,a} = \wt f^*_{t,a} + \delta_{t,a}$, so $\delta_{t,a}$ is the bias of $\widetilde{f}_{t,a}$. Also, let $X_0$ be a sample from the context distribution $\DD$, drawn  independently from $\F_T$. 
We will consider each term separately in the right-hand side of Equation~\eqref{eq:regret_dec}. First, for $\widetilde{R}_T$, we have:
\begin{align*}
 \widetilde R_T &= 	\EEt{ \sum_{t=1}^T \sum_{a\in \A} (\pi_t(a|X_0) - \pi^*(a|X_0))\hloss_{t,a}(X_0)}\\
					   & = 	\EEt{ \EEcct{\sum_{t=1}^T \sum_{a\in \A} (\pi_t(a|X_0) - \pi^*(a|X_0))\hloss_{t,a}(X_0)}{X_0}} \\
					   & = \EEt{ \EEcct{\sum_{t=1}^T \sum_{a\in \A} (\pi_t(a|X_0) - \pi^*(a|X_0))(\iprod{\varphi(X_0)}{\wt f^*_{t,a}} +  \iprod{\varphi(\tX)}{\delta_{t,a}}  - b_{t,a}(\tX))}{X_0}} \\
					   & = \EEt{  \sum_{t=1}^T \sum_{a\in \A} (\pi_t(a|X_t) - \pi^*(a|X_t))(\iprod{\varphi(X_t)}{ f_{t,a}} +  \iprod{\varphi(X_t)}{\EE{\delta_{t,a}}}  - b_{t,a}(X_t))  }\\
					   & = \EEt{  \sum_{t=1}^T \sum_{a\in \A} (\pi_t(a|X_t) - \pi^*(a|X_t))(\ell_{t,a}(X_t) +  \iprod{\varphi(X_t)}{\EE{\delta_{t,a}}}  - b_{t,a}(X_t))  } ,
\end{align*}
where we used the independence of $X_t$ and $\hS_{t,a}$. Applying the same sequence of equations to $B^*_T$ and $B_T$, we get
\begin{align*}
	B^*_T  = \EEt{  \sum_{t=1}^T \sum_{a\in \A}  \pi^*(a|X_t)(  \iprod{\varphi(X_t)}{\EE{\delta_{t,a}}}  - b_{t,a}(X_t))}
\end{align*}
and 
\begin{align*}
	B_T  = \EEt{  \sum_{t=1}^T \sum_{a\in \A}  \pi_t(a|X_t)(  \iprod{\varphi(\tX)}{  b_{t,a}(X_t) -\EE{ \delta_{t,a}} } )}.
\end{align*}
The proof is concluded by collecting all terms together.
\qed

\subsection{Proof of Lemma~\ref{lem:effective_dim_MGR}}\label{appx:effective_dim_MGR}
For the first part, recall the construction of $\hSigma_{t,a}^+$ defined through the KGR procedure. Note that $\EEt{B_{k,a}} = \Sigma_{t,a}$, and  $\{ X(k)\}_{k=1}^M$ are independent, and recall the identity 
$\pa{I + \sum_{i=1}^M U_i } (I-U) = \pa{I-U^{M+1}}$ that holds for any Hermitian operator $U: \ell_2 \to \ell_2$. Applying this identity with 
$U = I - \Sigma_{t,a}$, we get
\begin{align*}
\EE{\widehat{\Sigma}^{+}_{t,a}} \Sigma_{t,a} &=  \pa{I + \sum_{i=1}^M \pa{I - \Sigma_{t,a}}^i} \Sigma_{t,a} 
 = \pa{I - \pa{I -  \Sigma_{t,a}}^{M+1}}.
\end{align*} 
Let $\{e_1, e_2, \dots \}$ be the canonical basis in $\ell_2$ and recall that $\trace{U} = \sum_{i=1}^{\infty} \iprod{e_i}{U e_i} $. 
Also introducing the notation $\text{tr}_n\pa{U} \pa{U} = \sum_{i=n}^{\infty} \iprod{e_i}{U e_i} $ for $n \in \mathbb{N}$, we observe that the following holds for each $t,a$:
\begin{align}\label{eq:trace_sta}
     \text{tr}_n\pa{\Sigma_{t,a}}  &= \text{tr}_n\pa{\EE{\pi_t(a|X_t) \varphi(X_t) \otimes \varphi(X_t)}} \nonumber \\
     & = \EE{\pi_t(a|X_t) \text{tr}_n\pa{ \varphi(X_t) \otimes \varphi(X_t)}} \nonumber \\
     & \quad \quad\text{(by Fubini's theorem)}\nonumber\\
     & = \EE{\pi_t(a|X_t) \sum_{i=n}^{\infty} \iprod{e_i}{\varphi(X_t) \otimes \varphi(X_t) e_i} } \nonumber \\
     & = \EE{\pi_t(a|X_t) \sum_{i=n}^{\infty} \pa{\iprod{e_i}{\varphi(X_t)}}^2}  \nonumber \\
     & \le \EE{\pi_t(a|X_t) \sum_{i=n}^{\infty} \norm{\varphi(X_t)}_2^2 \norm{e_i}_2^2}  \nonumber \\
    & \quad\quad\text{(Cauchy--Schwarz inequality)} \nonumber\\
     & = \EE{\pi_t(a|X_t) \pa{\sum_{i=n}^{\infty}  \varphi_{i}^2(X_t)}} \nonumber\\
    & \quad\quad \text{(using $\norm{e_i}_2^2 = 1$)}\nonumber\\
    & = \EE{\pi_t(a|X_t) \pa{\sum_{i=n}^{\infty} \mu_i \psi_{i}^2(X_t)}} \le  \sum_{i=n}^{\infty} \mu_i,
\end{align}
where we used  $|\psi_i (x)|\le 1$ and $\mu_i\ge 0$ in the last step. 
Moving on to bounding the trace of $\Sigma_{t,a}\hSigma_{t,a}^+$, we observe
\begin{align*}
	\trace{\EE{\widehat{\Sigma}^{+}_{t,a}}\Sigma_{t,a}} &= \trace{ (I - (I - \Sigma_{t,a})^M)} = \sum_{i=1}^\infty \iprod{e_i}{\pa{I - (I-\Sigma_{t,a})^M} \cdot e_i }\\
 & \le m + \sum_{i=m+1}^\infty \iprod{e_i}{\pa{I - (I-\Sigma_{t,a})^M} \cdot e_i }\\
  & = m + \sum_{i=m+1}^\infty \pa{1 - ( 1- \iprod{e_i}{\Sigma_{t,a} \cdot e_i })^M} \\
 &\le m + M \sum_{i=m+1}^\infty\iprod{e_i}{\Sigma_{t,a} \cdot e_i}  = m + M \text{tr}_m(\Sigma_{t,a}) \\
	 &\le m + M \sum_{i=m+1}^\infty \mu_i,
	\end{align*}
where we have split the sum at some arbitrary $m$, used $\iprod{e_i}{\pa{I - (I-\Sigma_{t,a})^M} \cdot e_i } \le \norm{e_i}^2_2\le 1$ for the first $m$ terms, and used the 
inequality $1-M\lambda \le (1-\lambda)^M$ that holds for all $\lambda$ and $M > 1$ for the rest of the terms. Finally, we used used the inequality (\ref{eq:trace_sta}) in the last step.
The 
statement then follows from the taking  $m =m(\varepsilon)$.
\qed

\subsection{Proof of Lemma~\ref{lem:trace_M}}\label{appx:trace_M}
\begin{proof}
We start by writing the bias as
	\begin{align*} 
	\EEt{ \iprod{\varphi(x)}{ f_{t,a} } - \iprod{\varphi(x)}{ \wt{f}_{t,a}}}  
	&=   
	 \iprod{\varphi(x)}{ f_{t,a} }   - 
		\EEt{ \iprod{ \varphi(x)}{ \widehat{\Sigma}^{+}_{t,a}\varphi(X_t) } \iprod{ \varphi(X_t)}{ f_{t,a}} \II{A_t = 
					a}}
\\
&=   
	\iprod{\varphi(x)}{ f_{t,a} }   - 
		\EEt{ \iprod{ \varphi(x) }{\widehat{\Sigma}^{+}_{t,a}\Sigma_{t,a}f_{t,a} } }
\\
		& =  \EEt{ \iprod{\varphi(x)}{ \pa{ I - \widehat{\Sigma}^{+}_{t,a}\Sigma_{t,a}}  f_{t,a}} }
		\\
		& =  \iprod{\varphi(x)}{\pa{ \pa{I -  \Sigma_{t,a}}^{M+1}}  f_{t,a}}
		\\
		&\le 
		\alpha \norm{\varphi(x)}^2_{\pa{I -  \Sigma_{t,a}}^{M+1}} + \frac {1}{\alpha} \norm{f_{t,a}}^2_{\pa{I -  
\Sigma_{t,a}}^{M+1}},
\end{align*}
for an arbitrary $\alpha >0$,
where in the fourth line we have used the expression of $\EEt{\hSigma_{t,a}^+\Sigma_{t,a}}$ stated in Lemma~\ref{lem:trace_M}, and the Cauchy--Schwarz inequality in the last step.
The last term can be conveniently upper bounded by $\frac {1}{\alpha} \norm{f_{t,a}}^2 \le \frac{1}{\alpha}$. To bound 
the first term, notice that
\begin{align*}
 \alpha \norm{\varphi(x)}^2_{\pa{I -  \Sigma_{t,a}}^{M+1}} \le \frac{\alpha}{M+1}\sum_{i=0}^M 
\norm{\varphi(x)}^2_{\pa{I -  \Sigma_{t,a}}^{i}} = \frac{\alpha}{\pa{M+1}}
\EEt{\norm{\varphi(x)}^2_{\hSigma_{t,a}^+} },
\end{align*}
where we have used that $(I-\Sigma_{t,a})^{M+1} \preccurlyeq (I-\Sigma_{t,a})^{i}$ holds for all $i\le M+1$ due to $\opnorm{\Sigma_{t,a}}\le 1$, and the definition of $\hSigma_{t,a}^+$.
Putting the above statements together, we obtain
\begin{align*}
\EEt{ \iprod{\varphi(x)}{ f_{t,a} } - \iprod{\varphi(x)}{ \wt{f}_{t,a}}}  
	&\le \frac{\alpha}{ \pa{M+1}}\EEt{\norm{\varphi(x)}^2_{\hSigma_{t,a}^+} } + \frac{1}{\alpha}.
\end{align*}
The statement is then proved by taking $\alpha = \beta \pa{M+1}$.
\end{proof}

\subsection{Proof of Lemma~\ref{l:variance}}\label{appx:variance}
	The proof follows the steps of the proof of Lemma 6 of \citep{NO20}.
	We start by plugging in the definition of $\wt{f}_{t,a}$ and writing
	\begin{align*}
	\EEt{\sum_{a=1}^K    \pi_t(a|\tX)   \biprod{\varphi(\tX)}{\wt{f}_{t,a}}^2} &= \EEt{\sum_{a = 1}^K \pi_t(a|\tX)   
\pa{\iprod{\varphi(\tX)}{\hSp_{t,a} \varphi(X_t)}\cdot\iprod{ \varphi(X_t)}{f_{t,a} }\II{A_t=a}}^2} 
	\\
     &\le \EEt{\sum_{a = 1}^K \pi_t(a|\tX)   
    \pa{\iprod{\varphi(\tX)}{\hSp_{t,a} \varphi(X_t)}\II{A_t=a}}^2}
     \\
	&= \EEt{\sum_{a = 1}^K  \trace{\Sigma_{t,a} \hSp_{t,a} \Sigma_{t,a} \hSp_{t,a} 
		}},
	\end{align*}
     where we used $\biprod{\varphi(\tX)}{f_{t,a}} \le 1$ in the inequality.
     We omit $t,a$ indexes in the following text. 
	Using the Araki--Lieb--Thirring inequality, we get 
	\[  \trace{\Sigma \Sigma^+ \Sigma \Sigma^+} \le \trace{\Sigma^2 \pa{\Sigma^+}^2 }.\]

	Define $G_k = \pa{I -  B_k}$. Using the definition of 
	$\Sp$ and elementary manipulations, we can get 
	\begin{align*}
		\pa{\Sp}^2& = \pa{ I +  \sum_{k=1}^M \prod_{j=1}^k G_i}^2 = I+2\sum_{k=1}^{M}\prod_{j=1}^k G_j + 
\sum_{k,k'=1}^{M} \pa{\prod_{j=1}^k G_j }\pa{\prod_{j=0}^{k'} G_j }\\
		& = I + 2\sum_{k=1}^{M}\prod_{j=1}^k G_j + 2 \sum_{k=1}^{M}\sum_{k'=k}^{M} \prod_{j=1}^k G^2_j \prod_{j=k+1}^{k'} G_j - \sum_{k=1}^{M} \prod_{j=1}^k G^2_j\\
		& \preccurlyeq 2I + 2 \sum_{k=1}^{M}\prod_{j=1}^k G_j  + 2 \sum_{k=1}^{M}\sum_{k'=k}^{M} \prod_{j=1}^k G^2_j \prod_{j=k+1}^{k'} G_j,
	\end{align*}
	where in the second line we reordered the sum $\sum^M_{k,k' = 1} a_k a_{k'} = 2 \sum_{k=1}^{M} \sum_{k'=k}^{M}a_k a_{k'} - \sum_{k=1}^{M} a^2_k$, while in the third line we dropped the last term and added $I$. Denote $D=\EEt{ G_j}$ and $E = \EEt{G^2_j}$. Using independence of $G_j$'s we get:
	\[ \EEt{\Spt} \preccurlyeq 2 \sum_{k=0}^{M} D^k +2 \sum_{k=1}^{M} E^k \sum_{k'=0}^{M-k}D^{k'}.\]
	Using the fact that $D=I- \Sigma$, we have $\sum_{k=0}^{M}D^k \Sigma = I - D^M$ and thus
	\begin{align*}
 \EEt{\Spt  }\Sigma^2 &\preccurlyeq 2 \pa{I - D^M}\Sigma + 2\sum_{k=1}^{M} E^k 
\pa{I-D^{M-k}}\Sigma
\preccurlyeq 2 \Sigma + 2\sum_{k=1}^{M} E^k \Sigma,
	\end{align*}
	where we have also used the fact that if $A\preccurlyeq B$, then for any positive semi-definite operator 
$C$ holds the inequality $\trace{CA}\le \trace{CB}$.
Furthermore, since we have $ B  \preccurlyeq I$, we can also simplify
	\[ E = \EEt{(I-B)^2}  \preccurlyeq \EEt{(I- B)} = D, \]
	and write 
	\[
	 \sum_{k=1}^{M} E^k \Sigma \preccurlyeq  \sum_{k=1}^{M} D^k 
\Sigma = (I - (I - \Sigma)^M).
	\]
This then gives
	\begin{align*}
		\trace{\Sigma^2 \EEt{\Spt} } &\le  2 \trace{\Sigma} + 2 \sum_{k=1}^{M}\trace{ D^k 
\Sigma} =  2\trace{\Sigma}+ 2 \trace{ (I - (I - \Sigma)^M)}
\\
&\le 2 + m(\varepsilon) + M\varepsilon,
	\end{align*}
	where the last step follows from using Lemma~\ref{lem:effective_dim_MGR}.
\qed



\section{Implementing \kerlinexp}\label{appx:MGR}
We now present a fully operational definition of \kerlinexp that does away with all the abstractions used in the main 
text. In particular, we here provide a version that fully unpacks the computation of the cumulative loss estimates 
$\hL_{t}(X_t,a)$ needed by the FTRL subroutine to calculate the policy $\pi_t(\cdot|X_t)$. As some pondering of 
the abstract description reveals, this computation requires rerunning the entire KGR subroutine for the whole sequence 
of past observations, including reusing the context-action pairs generated by KGR in each time step preceding $t$. In 
order to accommodate this sample reuse, in Algorithm~\ref{alg:kerlinexp3_full}, we present a version of \kerlinexp that 
uses the subroutine \lossestimate to compute the cumulative loss estimates and a \emph{data buffer} $\BB_t$ that stores 
all relevant data for computing said estimates. This subroutine is presented as Algorithm~\ref{alg:lossestimate}, and 
it makes use of the KGR subroutine presented as Algorithm~\ref{alg:kgr}.

\begin{algorithm}
	\caption{\kerlinexp}
	\label{alg:kerlinexp3_full}
	\textbf{Parameters:} Learning rate $\eta>0$.\\
	\textbf{Initialization:} Set $\BB_t = \emptyset$.
	\\
	\textbf{For} $t = 1, \dots, T$, \textbf{repeat:}
	\begin{enumerate}[noitemsep]
		\item Observe $X_t$ and, for all $a$, compute $L_t(X_t,a) = \lossestimate(X_t,a, \BB_t)$.
		\item Observe $X_t$ and, for all $a$, set 
		\begin{equation}\label{eq:FTRL}
			\pi_{t}(\cdot|X_t) = \argmin_{p \in \Delta(\A)}\pa{\Psi(p) + \eta\sum_a p_a \widehat{L}_{t-1}(X_t,a) }, 
		\end{equation}
		\item draw $A_t$ from the policy $\pi_t(\cdot | X_t)$,
		\item observe the loss $\loss_t(X_t,A_t)$,
		\item for $k=1,2,\dots,M$, draw $X(k)\sim \mathcal{D}$ and $A(k) \sim \pi_t(\cdot|X(k))$ using the procedure above,
		\item update buffer with the tuple
		\[
		 \BB_{t+1} = \BB_t \cup \bpa{X_t, A_t, \ell_t(X_t,A_t), \ev{X_t(k),A_t(k)}_{k=1}^M}.
		\]
	\end{enumerate}
\end{algorithm}
\begin{algorithm}
 \caption{Kernel Geometric Resampling (\KGR)}\label{alg:kgr}
 \textbf{Parameters:} $\beta \ge 0$.\\
 \textbf{Input:} Context-action pairs $(x,a)$, $(x',a')$, and $\ev{x(k),a(k)}_{k=1}^M$.\\
			\textbf{For $k = 1, \dots, M$, repeat}:
			\begin{enumerate}[noitemsep]
				\vspace{-2mm}
				\item set  $B_{k,a} =  \II{a(k)=a}\varphi(x(k))\otimes\varphi(x(k))$,
				\item set $C_{k,a} = \prod_{j=1}^k\pa{I-  B_{j,a}}$,
				\item compute $q_{k,a}(x)= \iprod{\varphi(x)}{C_{k,a} \varphi(X_t)}$, and
				\item compute $b_{k,a}(x) = \beta \iprod{\varphi(x)}{C_{k,a} \varphi(x)}$.
			\end{enumerate}
			\vspace{-2mm}
			\textbf{Return $q(x,a) =  \II{a = a'}\pa{\kappa(x, x') +  \sum_{k=1}^M q_{k,a}(x)},$}
	
			$\	\qquad \quad b(x,a) = \kappa(x, x) + \sum_{k=1}^M b_{k,a}(x)$.
\end{algorithm}
%
%
%

\begin{algorithm}
 \caption{\lossestimate}\label{alg:lossestimate}
 \textbf{Input:} context $x$, action $a$, a set of tuples $\pa{X_i, A_i, \ell_i(X_i,A_i), 
\ev{X_i(k),A_i(k)}_{k=1}^M}_{i=1}^n$.\\
 \textbf{Initialize:} $L_0(x,a) = 0$ for all $a$.\\
 \textbf{For $i=1,2,\dots,n$, repeat:}
  \begin{enumerate} [noitemsep]
   \item let $(q_i(x,a),b_i(x,a)) = \KGR(x,a,X_i,A_i,\ev{X_i(k),A_i(k)}_{k=1}^M)$,
   \item let $\widehat\ell_i(x,a)= q_{i}(x,a) \loss_i(X_i, a) \II{A_i = a} - b_{i}(x,a)$,
   \item update $\hL_i(x,a) = \hL_{i-1}(x,a) + \hloss_i	(x,a)$.
  \end{enumerate}
\end{algorithm}

As we show in Appendix~\ref{app:computation}, the KGR procedure with a given set of inputs runs in $\OO(M^2)$ time. In round $t$, the KGR subroutine is called by \lossestimate $t$ times, which costs a total of $\OO(tM^2)$ time. Finally, \lossestimate is called by the main algorithm \kerlinexp $M+1$ times when generating the action $A_t$ and the independent copies $\ev{A_t(k)}_{k=1}^M$, which altogether makes for a time complexity of $\OO(tM^3)$ per round. Thus, the total time complexity of implementing \kerlinexp is $\OO(T^2 M^3)$.
As for memory complexity, the main bottleneck is having to store the data buffer $\BB_t$, which consists of $t(M+1)$ context-action pairs and $t$ observed losses. Overall, this means that \kerlinexp requires to store a total of $\OO(TM)$ context-action pairs in memory.

\section{Computational analysis of KGR}\label{app:computation}
\begin{lemma}
 Kernel Geometric Resampling requires $\OO(M^2)$ elementary operations and requires $\OO(M)$ memory.
\end{lemma}
\begin{proof}
The proof goes by induction. For $M=1$, we get
\begin{align*}
    q_{t}(x,a) &= k(x, X_t) + \iprod{\varphi(x)}{\pa{I - \II{a(1)=a}\varphi(X(1))\otimes\varphi(X(1))  } \varphi(x)}\\
    &= 2k(x, X_t) -  \II{A(1)=a} \kappa(x, X(1))k(X(1), X_t).
\end{align*}
For step $m+1$, we have 
\begin{align*}
    q_{t,k, a}(x)&= \siprod{\varphi(x)}{\prod_{j=1}^k\pa{I- B_{j,a}} \varphi(X_t)}
    \\
    &= \siprod{\varphi(x)}{\prod_{j=1}^{k-1}\pa{I- B_{j,a}} \cdot \pa{I- 
\II{a(k)=a}\varphi(X(k))\otimes\varphi(X(k))}  \varphi(X_t)} \\
    & =  q_{t,k-1, a} + \II{a(k)=a} \siprod{\varphi(x)}{\prod_{j=1}^{k-1}\pa{I- B_{j,a}} \cdot 
\varphi(X(k))\otimes \varphi(X(k))   \varphi(X_t)}.
\end{align*}
Note that there exists a set of coefficients $ \{ p_{a,k-1, i} \}_{i=1}^{k-1}$ such that 
\begin{align*}
    \siprod{\varphi(x)}{
\prod_{j=1}^{k-1}\pa{I- B_{j,a}}  \cdot 
  \varphi(X_t) } = \sum_{i= 1}^{k-1} p_{a,k-1, i} \iprod{\varphi(X(i))}{ \varphi(X_t)  }.
\end{align*}
We can compute $ \{ p_{a,k, 
i} \}_{i=1}^{k}$ in $k$ steps, as:
\begin{align*}
   & \siprod{\varphi(x)}{ \prod_{j=1}^{k}\pa{I- B_{j,a}}  \cdot 
   \varphi(X_t) } \\
  &= \sum_{i= 1}^{k} p_{a,k-1, i} \iprod{\varphi(X(i)) }{  \pa{I- B_{k,a}} \cdot 
  \varphi(X_t) }   \\
    &= \sum_{i= 1}^{k} p_{a,k-1, i} \iprod{\varphi(X(i)) }{   \varphi(X_t) } \\
    &\qquad \qquad  - \II{a(k)=a} \sum_{i= 1}^{k} p_{a,k-1, i} \iprod{\varphi(X(i))}{ \varphi(X(k+1)) \otimes \varphi(X(k+1)) \varphi(X_t) } \\
    & = \sum_{i= 1}^{k} p_{a,k-1, i} \iprod{\varphi(X(i)) }{\varphi(X_t)} \\
    & \qquad \qquad  - \II{a(k)=a} \sum_{i= 1}^{k} p_{a,k-1, i} \iprod{\varphi(X(i))}{ \varphi(X(k+1))} \iprod{ \varphi(X(k+1))}{ \varphi(X_t) } \\
    & = \sum_{i= 1}^{k} p_{a,k-1, i} \iprod{\varphi(X(i)) }{\varphi(X_t)} \\
    & \qquad  \qquad  - \II{a(k)=a} \sum_{i= 1}^{k} p_{a,k-1, i} \kappa(X(i), X(k+1))\iprod{ \varphi(X(k+1))}{ \varphi(X_t) }. 
\end{align*} 
Thus, we get that for $i \le k-1$, $ p_{a,k, i} =  p_{a,k-1, i}$ and $p_{a,k, k}  = - \II{a(k)=a} \sum_{i= 
1}^{k-1} p_{a,k-1, i} \kappa((X(i), X(k))  $, which means that computing $p_{a,k, i}$ takes $k-1$ operations, which results in $\frac{M(M+1)}{2}$ operations to compute $ \{ 
p_{a,k-1, i} \}_{i=1}^{M}$.
In order to do this we need to store in memory an array of size $M$ with coefficients $ \{ 
p_{a,k-1, i} \}_{i=1}^{k-1}$. Notice, that the same line of computations apply to computing $b_{t,a}(x)$. Thus, given that the time of computing kernel is $K$, to compute $q_{t,a}(x)$ and $b_{t,a}(x)$, we need 
$\frac{M(M+1)}{2} K$ steps. 
\end{proof}

\end{document}